\documentclass[sigconf, nonacm, pdfa=true]{acmart}
\renewcommand\footnotetextcopyrightpermission[1]{}
\AtBeginDocument{%
  }

\usepackage{romannum}

\usepackage{float}
\usepackage{subcaption}  
\usepackage{tikz}
\usepackage{pgfplots}
\pgfplotsset{compat=1.18}

\usepackage{balance}
\balance

\begin{document}

\title{GuideTouch: An Obstacle Avoidance Device with Tactile Feedback for Visually Impaired }



\author{Timofei Kozlov, Artem Trandofilov*, Georgii Gazaryan*, \\
Issatay Tokmurziyev, Miguel Altamirano Cabrera, Dzmitry Tsetserukou}

\thanks{*These authors contributed equally to this work.}

\affiliation{%
  \institution{Skolkovo Institute of Science and Technology (Skoltech)}
  \city{Moscow}
  \country{Russia}
}
\email{{Timofei.Kozlov, Artem.Trandofilov, Georgii.Gazaryan, Issatay.Tokmurziyev, m.altamirano, d.tsetserukou}@skoltech.ru}

\renewcommand{\shortauthors}{Kozlov et al.}

\begin{abstract}

Safe navigation for the visually impaired individuals remains a critical challenge, especially concerning head-level obstacles, which traditional mobility aids often fail to detect. We introduce GuideTouch, a compact, affordable, standalone wearable device designed for autonomous obstacle avoidance. The system integrates two vertically aligned Time-of-Flight (ToF) sensors, enabling three-dimensional environmental perception, and four vibrotactile actuators that provide directional haptic feedback.  Proximity and direction information is communicated via an intuitive 4-point vibrotactile feedback system located across the user's shoulders and upper chest. For real-world robustness, the device includes a unique centrifugal self-cleaning optical cover mechanism and a sound alarm system for location if the device is dropped. We evaluated the haptic perception accuracy across 22 participants (17 male and 5 female, aged 21–-48; mean 25.7, $\pm$ 6.1). Statistical analysis confirmed a significant difference between the perception accuracy of different patterns. The system demonstrated high recognition accuracy, achieving an average of 92.9\% for single and double motor (primary directional) patterns. Furthermore, preliminary experiments with 14 visually impaired users validated this interface, showing a recognition accuracy of 93.75\% for primary directional cues. The results demonstrate that GuideTouch enables intuitive spatial perception and could significantly improve the safety, confidence, and autonomy of users with visual impairments during independent navigation.

\end{abstract}



\begin{teaserfigure}
       \centering
  \includegraphics[width=1.0\textwidth]{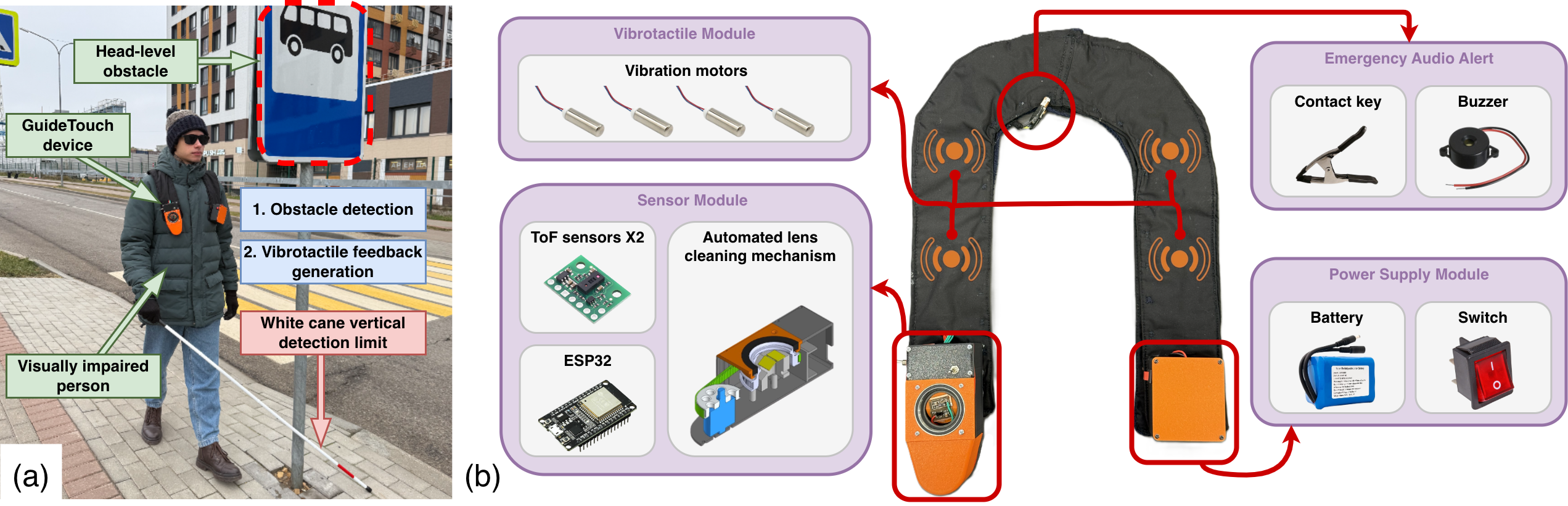}
  \caption{GuideTouch wearable obstacle avoidance system for visually impaired navigation. (a) Usage scenario with the device worn during outdoor navigation. (b) System architecture comprising a vibrotactile module, a sensor module, an auditory alert module, and a power supply.}
  \Description{This figure shows the architecture and use scenario of the GuideTouch -  wearable obstacle avoidance device for the visually impaired. On the left, a user who is blind walks outdoors near an obstacle at head level, which cannot be reached by a white cane. The obstacle is detected by infrared ToF sensors on the GuideTouch device, worn on the shoulders, and the directional vibrotactile feedback given to the user by four embedded vibration motors in the scarf-like wearable allows the user to avoid the obstacle before physical contact. On the right (b), a hardware diagram breaks down the main functional blocks of the GuideTouch device. The Vibrotactile Feedback Block includes four vibration motors placed on the shoulders and upper chest for spatially distributed alerts. The Sensor Block consists of two ToF sensors that measure distance, an ESP32 microcontroller that performs the processing, and a self-cleaning optical cover mechanism to maintain sensor clarity in harsh outdoor conditions. The Sound Alarm System consists of a buzzer and a contact key to help locate the device if it falls off. Finally, the Power Supply Block consists of a rechargeable battery and an external power switch for user control. Each component is physically integrated into the wearable structure, as shown on the actual photo of the device on the right.}
  \label{fig:teaser}
\end{teaserfigure}

\maketitle

\section{Introduction}
Visual impairment substantially reduces personal autonomy and mobility. According to the WHO, around 295 million people are visually impaired and 43 million are blind worldwide \cite{WHO23}. Navigation in dynamic environments remains difficult: more than 40\% of blind individuals report head‑level injuries or falls every few months, regardless of mobility aids \cite{UCSC10}. Modern assistive navigation technologies increasingly combine real‑time obstacle detection with haptic feedback \cite{Xu23_WearableETAReview, Joseph23_WearableETAReview, Eagleman23_HapticDevices}, which offers a silent and intuitive orientation channel compared to auditory cues that may be masked by environmental noise. Studies confirm that tactile feedback improves safety by providing immediate, directionally relevant information through wearable devices.

Existing solutions \cite{Katzschmann18_ToFHaptic, Skulimowski25_HapticAuditoryNavigation, Leporini22_HapticWearable, Bazhenov24_DogSurf} range from deep‑learning smart glasses and VLM-based interaction systems \cite{Tokmurziyev25_LLMGlasses, Gao25_WOAD, Khan25_HapticVLM} to wearable ToF systems, but all exhibit limitations. Computer Vision based models such as YOLO require substantial computational power \cite{Said23_DLObstacle, Xu23_HeadMountedObstacle}, making devices bulky or dependent on external processing. Smart-glasses based designs may be uncomfortable or culturally unsuitable, while camera systems degrade in low‑light conditions \cite{Elmannai17_SensorAssistive}. Several solutions also assume replacing rather than complementing the white cane, which our user study indicates is highly undesirable.

We propose a wearable device that detects obstacles via infrared Time‑of‑Flight sensors and provides haptic feedback through four vibration motors positioned on the shoulders and upper chest \cite{Park25_FullBodyHaptics, VibroForearm23}. The system includes automatic optical cover cleaning and a location alarm in case the device is dropped. 

In Section~\ref{sec:system}, the architecture of the system is described with subsections that describe each subsystem in more detail. In Section ~\ref{sec:experiment}, the experiment is conducted to validate the accuracy of haptic perception of several vibrotactile patterns.

To validate system requirements, we interviewed 28 visually impaired participants (17 male, 11 female, aged 16–-60). They identified three major categories of undetected obstacles: head‑level hazards (road signs, gates, horizontal bars), drop‑offs and ground‑level risks (open manholes, stairs), and thin objects (poles, bollards, fencing). Participants expressed strong interest in a hands‑free, wearable device capable of covering these blind spots—especially head‑level obstacles, which were reported as the most frequent and dangerous.

\section{System Architecture}\label{sec:system}

GuideTouch is an obstacle avoidance device based on ToF sensors, which produce a depth image of the area in front of the wearer. The device analyzes the image using an ESP32 microcontroller. The general scheme of the prototype is shown in Fig.~\ref{fig:teaser}. The system consists of two boxes and a scarf that holds them together, with a total weight not exceeding 500 g, and an estimated cost of approximately 100 USD. Vibration motors, the alarm system, an electronic speed controller, and wiring are installed on the inside of the scarf. One of the boxes contains a battery and a general power switch, providing up to 12 hours of autonomous operation, or 4 hours while rotating the optical cover. The other box includes a microcontroller ESP32, two ToF sensors, and a self-cleaning optical cover mechanism, which consists of a brushless direct current (BLDC) motor and mechanical transmission. The switch for turning on the mechanism is located on the side of the second box.

\subsection {Mechanics}

Raindrops and snowflakes pose a significant problem for wearable devices that rely on cameras or other light-based sensors. We investigated several approaches for cleaning the optical cover that is in front of the ToF sensors: an ultrasonic membrane, which breaks droplets into smaller particles and detaches them from the surface; vibration motors, which help droplets flow down more quickly; and a spinning mechanism that relies on centrifugal force, using a BLDC motor and a belt drive to rotate the optical cover (in the form of circular infrared transparent glass) at 3000 rpm.

During our testing, the spinning mechanism demonstrated the best performance, effectively removing droplets of any size in 18 out of 20 cases and leaving no visible traces, whereas the other methods failed to clear all droplets or removed them too slowly. The noise level generated by this chosen cleaning system measured an average of 70 dB at head level, a sound level considered safe for continuous exposure exceeding the mechanism’s expected operating time \cite{Fink2017_SafeNoiseLevel}.

\subsection{Electronics and Sensors}    

\begin{figure}[H]
    \centering
    \includegraphics[width=0.7\linewidth]{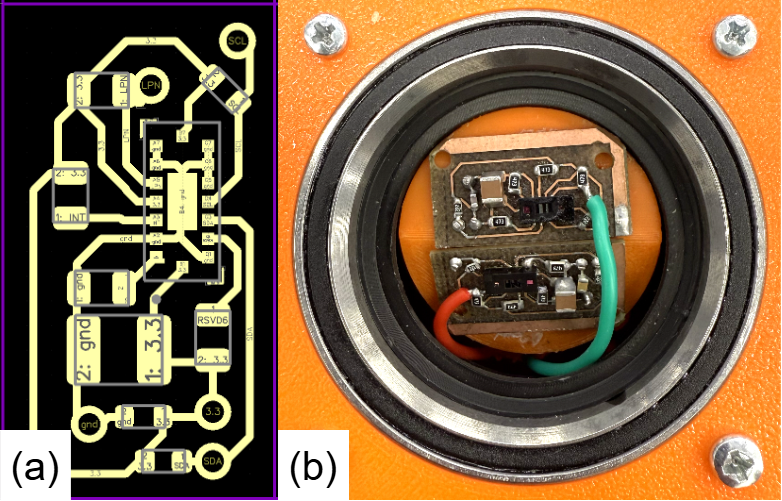}
    \caption{Custom ToF sensor module. (a) PCB layout design, (b) Assembled board with dual VL53L5CX sensors mounted behind the optical lens.}
    \Description{Figure shows the sensor hardware in the GuideTouch system.
Image (a) shows the designed layout of a custom PCB, intended for connecting a ToF distance sensor. Image (b) is a close-up photo of the assembled sensor module embedded inside the GuideTouch device. Two stacked PCBs are attached: the first is aiming straight ahead, while the second one aims a bit down. The whole assembly is fixed behind the circular optical cover and enclosed in the mechanical self-cleaning module. }
    \label{fig:sensor_latest}
    \vspace*{-0.4cm}
\end{figure}
GuideTouch uses two vertically aligned ToF multizone ranging sensors (VL53L5CX) on custom printed circuit boards (PCBs) to obtain distance data (Fig.~\ref{fig:sensor_latest}). Each sensor provides an 8x8 matrix of distances and has a $65^{\circ}$ diagonal square Field of View. The two sensors are positioned at a $30^{\circ}$ angle relative to each other, achieving a combined vertical Field of View of $90^{\circ}$. This configuration allows the device, when worn by a user of average height (170 cm), to detect obstacles at knee level (30 cm) and head level (160 cm) from a distance of 50 cm. The size of a detectable obstacle can be calculated based on the sensor's view angle and the number of measuring zones. The linear size $s$ of the zones can be calculated with the following formula:
\begin{equation}
s = 2d \cdot \tan\left( \frac{1}{2} \cdot \frac{\text{FoV}}{N} \cdot \frac{\pi}{180} \right), \label{eq:cell_size}
\end{equation}
where ${d}$ is the distance from the user to the obstacle, ${FoV}=60^\circ$ is the sensor's view angle, ${N}=8$ is the number of zones on each side of the view square. Sensors can detect obstacles that occupy approximately 30\% of a view zone's area. At a distance of 1 m the device is able to detect obstacles with the linear size of 4 cm.

The Alarm system consists of a passive piezo buzzer with a timer module NE555 and an electrical contact key (clip). The clip is placed on a person's collar or other part of clothing. If the device becomes dislodged, the clip closes the circuit, and the buzzer produces a high-pitched noise (chosen around 3--4 kHz) to notify the user of the device's position.

\subsection{Haptic Feedback and Navigation}

\begin{figure}[H]
    \centering
    \includegraphics[width=1.0\linewidth]{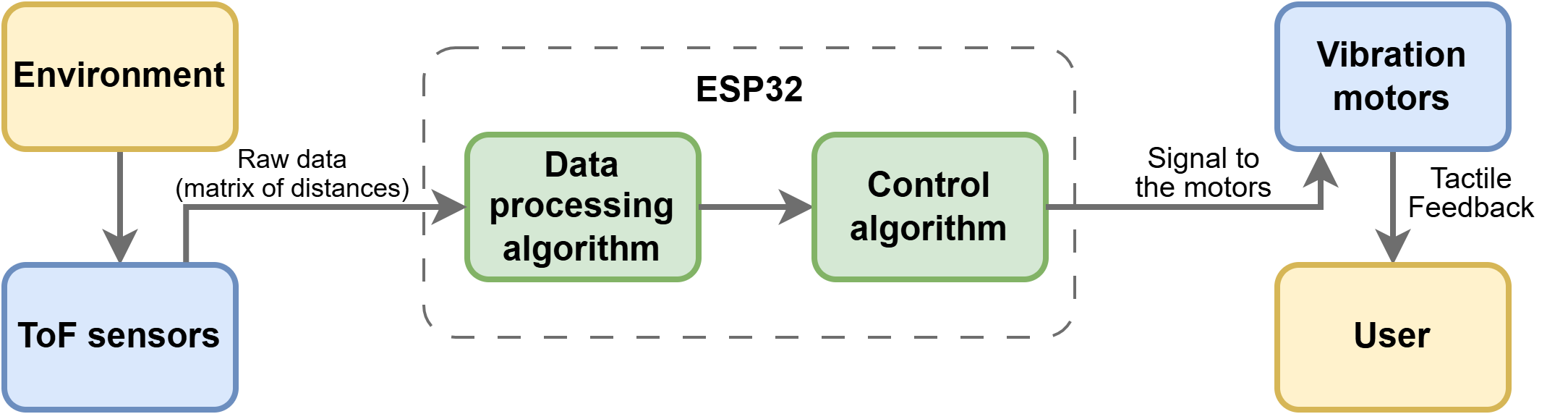}
    \caption{System operational workflow.}
    \Description{The figure shows the flow diagram of data in GuideTouch, where on the left, the environment is sensed using Time-of-Flight sensors that produce a raw matrix of distance measurements. Data is then sent to the ESP32 microcontroller, which has a custom algorithm to process this matrix. A control algorithm then uses this processed information to generate signals that turn on vibration motors, which are used to provide a user with vibrotactile feedback about the presence and direction of obstacles detected. The user receives feedback and, based on that, can adapt their movements.}
    \label{fig:dataflow}
    \vspace*{-0.4cm}
\end{figure}

\begin{figure}[H]
    \centering
    \includegraphics[width=1\linewidth]{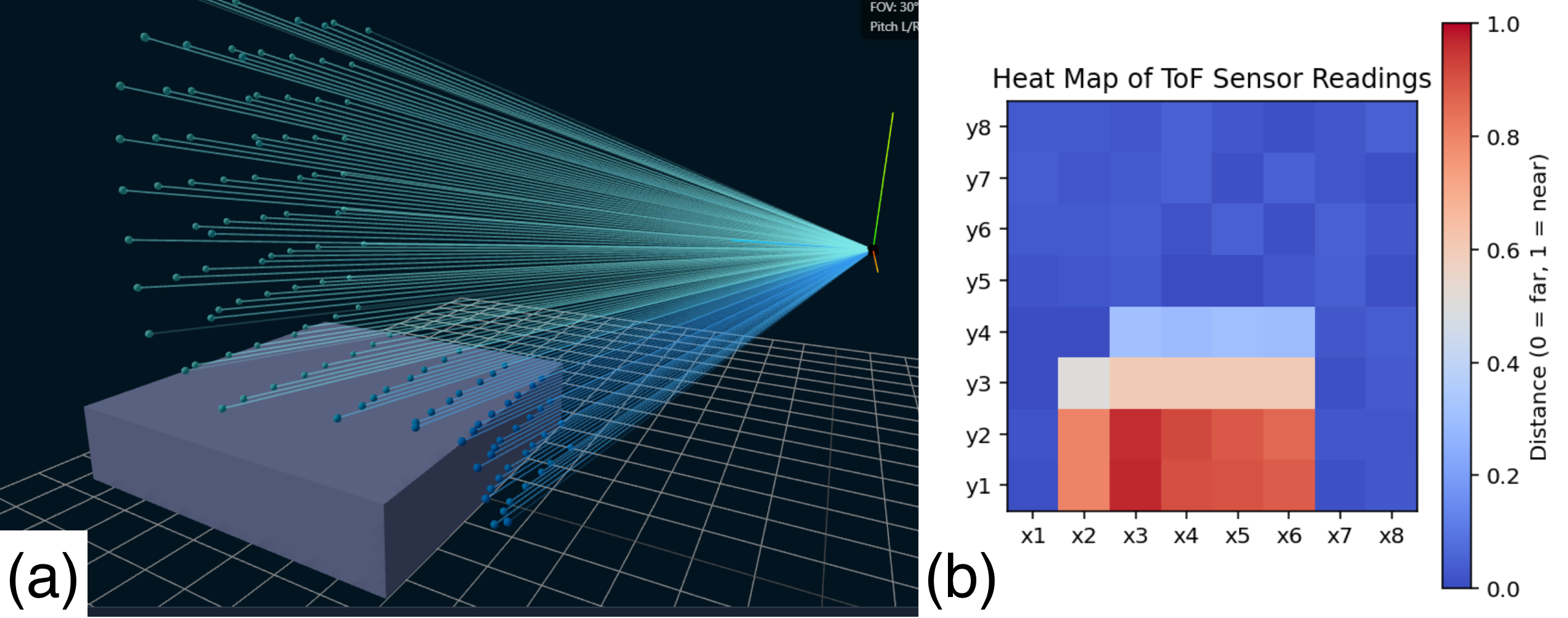}
    \caption{Sensor data visialization methods. (a) 3D point cloud of obstacle geometry. (b) 2D heat map of distance measurements.}
    \Description{The figure consists of two subfigures showing the visualisation of data obtained from Time-of-Flight sensors. Subfigure (a) shows a 3D rendering of distance vectors produced by a Time-Of-Flight sensor. Blue rays are cast towards a nearby rectangular object, with distance points visualised as small spheres on the object surface. Subfigure (b) presents a heat map of the sensor readings. The grid is coloured according to the proximity of objects: red indicates close distances (near obstacles), while blue represents farther regions. The image on the heat map corresponds to a Depth Image of a rectangular object}
    \label{fig:heat_plus_range}
    \vspace*{-0.4cm}
\end{figure}

Four vibration motors are embedded into the fabric of the scarf: two positioned above each enclosure, and two placed over the shoulders on either side of the neck. Each vibration motor corresponds to a specific direction. The general work process of the proposed device is visualized in Fig.~\ref{fig:dataflow}. The ESP32 microcontroller receives the distance data from the ToF sensors every 0.1 seconds. Then it processes these data, filters any noise using temporal outlier filtering, and analyzes the depth map for the presence of obstacles (Fig.~\ref{fig:heat_plus_range}). The algorithm separates the pictures into several zones (e.g., left, right, top) and produces the signals for vibration motors. The device vibrates on the side closest to the obstacle, informing the user of the direction of the possible danger. If an obstacle covers multiple zones (e.g., an overhanging section) the device activates the set of motors corresponding to all dangerous directions.

 The device functions as a human-oriented analog of a parking sensor system, providing proximity feedback through haptic cues. The device vibrates on the side that is closest to the obstacle ahead, thus informing the user of the direction of the possible danger. If the obstacle covers multiple zones (e.g., a sudden horizontal wall protrusion with an overhanging section, posing a serious risk of head impact), the device will vibrate with a set of vibration motors that correspond to all the dangerous directions.

\section{System Evaluation} \label{sec:experiment}
The system evaluation was conducted in two phases: assessing the accuracy of perception of the haptic interface and performing preliminary validation with visually impaired end-users.

\subsection{Participants}

To evaluate the performance of the haptic feedback subsystem and to analyze the precision of vibrotactile pattern recognition, we conducted a series of experiments involving 22 participants (17 male and 5 female, aged 21 to 48; mean 25.7, $\pm$ 6.1) without previous experience on the system. The participants were divided into two equal-sized groups, A and B, where group A experienced all possible vibrotactile patterns with single, double, and triple motor combinations (15 patterns) and group B experienced only single and double motor vibration (10 patterns).

\subsection{Procedure}

At the beginning of the experiment, the participant was invited to sit in front of a laptop and wear the GuideTouch device. Then a training session was conducted, in which the user became familiar with the vibrotactile patterns. After reading the initial instruction, the experiment coordinators turned on the GuideTouch device and motors started vibrating in a predetermined order. The experiment coordinators announced to the participant which motors were vibrating at that exact time. Group A experienced vibrotactile patterns that included 1, 2, and 3 motors. Group B experienced vibrotactile patterns that included only single and double motor patterns.

After the initial stage of the experiment, vibrations stopped and the experiment coordinators read the next part of the instructions. The Python script was launched on the laptop, which displayed the experiment GUI. The participant was instructed to familiarize themselves with the GUI, after which they pressed the Start button and the Python script began sending vibration commands to the device, starting the vibrations. The order of the vibrotactile patterns was randomized by the script, each pattern occurring 5 times during the experiment. Each participant in Group A experienced $5 \times 15 = 75$ vibrotactile patterns and each participant in Group B experienced $5 \times 10 = 50$ vibrotactile patterns. Each vibration lasted for 3 seconds, after or during which the participant chose the button in the GUI that corresponded to the vibrotactile pattern currently being active. After the participant's answer, the next vibration started until each pattern was repeated 5 times. The Python script recorded the vibration pattern that was sent to the device and executed as well as the participant's perceived pattern which they chose in the GUI. The total number of measurements is 825 for Group  A (15 patterns \(\times\) 5 repetitions \(\times\) 11 users) and 550 for Group B (10 patterns \(\times\) 5 repetitions \(\times\) 11 users).

\subsection{Results}

To visualize the results, confusion matrices for each group were created (Table~\ref{table:table_A} and Table~\ref{table:table_B}). Each vibration motor was assigned a corresponding abbreviation. The bottom left motor is L1, the bottom right motor is R1, the upper left motor is L2 and the upper right motor is R2. For abbreviating patterns that consist of combinations of motors, the `+' sign is used (e.g. the pattern consisting of the upper right and the bottom left motors is abbreviated as L1 + R2).

\begin{table*}[ht]
\centering
\caption{Confusion Matrix of Vibrotactile Patterns for Group A (11 participants), values in \%}
\label{tab:confusion_full}
\vspace{0.5em}
\resizebox{\textwidth}{!}{%
\begin{tabular}{l|cccc|cccccc|cccc|c}
\toprule
\textbf{True \textbackslash\ Pred.} & L1 & L2 & R1 & R2 & L1+L2 & R1+R2 & L1+R1 & L2+R2 & L1+R2 & L2+R1 & L1+L2+R1 & L1+L2+R2 & L1+R1+R2 & L2+R1+R2 & L1+L2+R1+R2 \\
\midrule
L1 & \textbf{98} & 2 & 0 & 0 & 0 & 0 & 0 & 0 & 0 & 0 & 0 & 0 & 0 & 0 & 0 \\
L2 & 2 & \textbf{93} & 0 & 0 & 0 & 0 & 0 & 0 & 0 & 0 & 2 & 2 & 0 & 2 & 0 \\
R1 & 0 & 0 & \textbf{96} & 2 & 0 & 0 & 0 & 2 & 0 & 0 & 0 & 0 & 0 & 0 & 0 \\
R2 & 0 & 0 & 0 & \textbf{98} & 0 & 2 & 0 & 0 & 0 & 0 & 0 & 0 & 0 & 0 & 0 \\
\midrule
L1+L2 & 5 & 9 & 0 & 0 & \textbf{85} & 0 & 0 & 0 & 0 & 0 & 0 & 0 & 0 & 0 & 0 \\
R1+R2 & 0 & 0 & 4 & 15 & 0 & \textbf{80} & 0 & 0 & 0 & 0 & 0 & 0 & 2 & 0 & 0 \\
L1+R1 & 2 & 0 & 0 & 0 & 0 & 0 & \textbf{96} & 0 & 0 & 0 & 2 & 0 & 0 & 0 & 0 \\
L2+R2 & 0 & 0 & 0 & 0 & 0 & 0 & 0 & \textbf{98} & 0 & 0 & 0 & 0 & 0 & 2 & 0 \\
L1+R2 & 0 & 0 & 0 & 0 & 0 & 0 & 2 & 0 & \textbf{73} & 4 & 0 & 9 & 7 & 5 & 0 \\
L2+R1 & 0 & 5 & 0 & 0 & 0 & 0 & 0 & 0 & 5 & \textbf{65} & 11 & 2 & 0 & 9 & 2 \\
\midrule
L1+L2+R1 & 0 & 0 & 0 & 0 & 2 & 0 & 7 & 2 & 2 & 5 & \textbf{65} & 5 & 2 & 0 & 9 \\
L1+L2+R2 & 0 & 2 & 0 & 0 & 0 & 0 & 2 & 16 & 2 & 0 & 4 & \textbf{64} & 0 & 4 & 7 \\
L1+R1+R2 & 0 & 0 & 0 & 0 & 0 & 2 & 13 & 0 & 18 & 0 & 4 & 5 & \textbf{42} & 11 & 5 \\
L2+R1+R2 & 0 & 2 & 0 & 0 & 0 & 0 & 0 & 20 & 0 & 5 & 2 & 4 & 9 & \textbf{56} & 2 \\
\midrule
L1+L2+R1+R2 & 0 & 0 & 0 & 0 & 0 & 0 & 2 & 7 & 0 & 0 & 13 & 2 & 2 & 9 & \textbf{65} \\
\bottomrule
\end{tabular}
}
\label{table:table_A}
\end{table*}

\begin{table}[ht]
\centering
\caption{Confusion Matrix of Vibrotactile Patterns For Group B (11 participants), values in \%}
\footnotesize
\setlength{\tabcolsep}{3pt}
\renewcommand{\arraystretch}{1.1}
\begin{tabular}{l|cccccccccc}
\toprule
\textbf{True \textbackslash\ Pred.} & L1 & L2 & R1 & R2 & L1+L2 & R1+R2 & L1+R1 & L2+R2 & L1+R2 & R1+L2 \\
\midrule
L1       & \textbf{98} & 2  & 0  & 0  & 0  & 0  & 0  & 0  & 0  & 0 \\
L2       & 0  & \textbf{95} & 0  & 4  & 0  & 0  & 0  & 2  & 0  & 0 \\
R1       & 0  & 0  & \textbf{95} & 4  & 0  & 0  & 0  & 2  & 0  & 0 \\
R2       & 0  & 0  & 0  & \textbf{98} & 0  & 0  & 0  & 0  & 2  & 0 \\
L1+L2    & 4  & 11 & 0  & 0  & \textbf{84} & 0  & 0  & 0  & 2  & 0 \\
R1+R2    & 0  & 0  & 4  & 11 & 0  & \textbf{80} & 2  & 2  & 0  & 2 \\
L1+R1    & 2  & 4  & 0  & 0  & 0  & 0  & \textbf{93} & 2  & 0  & 0 \\
L2+R2    & 0  & 0  & 0  & 2  & 0  & 0  & 0  & \textbf{98} & 0  & 0 \\
L1+R2    & 0  & 0  & 0  & 0  & 0  & 0  & 0  & 2  & \textbf{98} & 0 \\
R1+L2    & 0  & 5  & 0  & 2  & 0  & 0  & 0  & 2  & 0  & \textbf{91} \\
\bottomrule
\end{tabular}
\label{table:table_B}
\end{table}

The average accuracy for perceiving the vibrotactile patterns is 78.4\% for Group A and 92.9\% for Group B. The difference in average accuracy can be explained by the difference in the complexity of the patterns between two groups. This can be further observed in the Confusion Matrix for Group A. The accuracy for single and double motor patterns is higher than for 3 and 4 motor patterns, while the majority of errors in single and double motor patterns were caused by confusing them with 3 and 4 motor patterns.

For Group A (All Patterns), one-way repeated measurements ANOVA on vibration patterns revealed a statistically significant difference: $F(14,150)=7.5339,$ $p<0.000001$. Post-hoc analysis (Bonferroni and Tukey's HSD) confirmed that the significant differences were concentrated between simpler patterns (1 and 2 motors) and more complex patterns (3 and 4 motors), indicating that increased spatial overlap reduces recognition accuracy.

For Group B (Simple Patterns), the one-way repeated measurements ANOVA did not show statistically significant difference $F(9, 100) = 1.227,$ $p = 0.287$. However, standard one-way ANOVA on participants revealed a significant effect: $F(10, 99) = 3.8842,$  $p = 0.0002$. Tukey's HSD test detected statistically significant differences between two female participants and the rest. Notably, these two female participants had reported suboptimal contact with the device during the experiment, which they attributed to physiological discomfort, potentially affecting their recognition accuracy.

\subsection {Preliminary Experiments with Visually Impaired Participants}

We conducted preliminary experiments with blind participants, including visits to several associations for the blind and visually impaired support centers. During these tests, we evaluated the ability of blind users (14 participants, 8 male and 6 female, aged 16–-60)  to accurately identify vibrotactile patterns generated by our device. The results demonstrated an average recognition accuracy of 93.7\%, indicating the effectiveness of our approach. 

\section{Conclusion and Future Work}

In this paper, we introduced GuideTouch, a compact standalone wearable device designed to assist individuals with visual impairments in avoiding obstacles, particularly upper-body obstacles, which are a common cause of injury. The system combines the use of Time-of-Flight sensors for 3-D environmental perception with a 4-point tactile feedback system to convey spatial cues. The conducted experiments validated the utility of vibration motors as a practical way to convey directional information to the user. Statistical analysis showed a high average recognition accuracy (92.9\%) for single and double motor patterns. Post-hoc analysis indicated a significant variance in the perception of vibrotactile patterns based on their complexity, and a significant difference in accuracy between some participants, suggesting a dependency on wearability and potentially physiological differences. Preliminary experiments with blind participants further demonstrated a high recognition accuracy (93.75\%) for primary directional cues, confirming the effectiveness of the interface.

While these findings validate the interface concept, they were obtained with first-time users and under static conditions. Based on the results of the experiments, we formulated the hypothesis that with training — or when tested by blind individuals who are more sensitive to tactile stimuli — the recognition accuracy of vibrotactile patterns may improve \cite{Gao25_hapticTouchscreen, VibroMotorTrain22, Cuppone16_ProprioVibro, Stronks17_VibroBack}. Future work will focus on a comprehensive evaluation of the system in dynamic scenarios with multiple-day training periods for participants to heighten their perception of vibrotactile stimuli. In the immediate future, we plan to integrate a higher-resolution ToF sensor, which would be implemented in a Computer Vision system \cite{Kolb10_ToF} to help navigating in the crowds. At the same time, we plan to start training sessions with the participants to familiarize them with the vibrations.

\section*{Acknowledgements} 
Research reported in this publication was financially supported by the RSF grant No. 24-41-02039.

\end{document}